\documentclass[10pt,twocolumn,letterpaper]{article}

\usepackage{cvpr}
\usepackage{times}
\usepackage{epsfig}
\usepackage{graphicx}
\usepackage{amsmath}
\usepackage{amssymb}

\usepackage{authblk}
\usepackage{framed,multirow}
\usepackage{booktabs}
\usepackage{subfigure}
\usepackage{enumerate}
\usepackage{pbox}

\usepackage[breaklinks=true,bookmarks=false]{hyperref}

\cvprfinalcopy 

\def\cvprPaperID{4} 

\begin{document}

\title{IR2VI: Enhanced Night Environmental Perception by Unsupervised Thermal Image Translation}
\author[1]{Shuo Liu\thanks{luke6649@outlook.com}}
\author[2]{Vijay John\thanks{vijayjohn1983@gmail.com}}
\author[3]{Erik Blasch\thanks{erik.blasch@gmail.com}}
\author[1]{Zheng Liu\thanks{zheng.liu@ieee.org}}
\author[4]{Ying Huang\thanks{huangying@cqupt.edu.cn}}
\affil[1]{University of British Columbia, BC, Canada }
\affil[2]{Toyota Institute of Technology, Nagoya, Japan}
\affil[3]{Air Force Research Laboratory, VA, United States}
\affil[4]{Chongqing University of Posts and Telecommunications, Chongqing, China}

\maketitle
\begin{abstract}
 Context enhancement is critical for night vision (NV) applications, especially for the dark night situation without any artificial lights. In this paper, we present the infrared-to-visual (IR2VI) algorithm, a novel unsupervised thermal-to-visible image translation framework based on generative adversarial networks (GANs). IR2VI is able to learn the intrinsic characteristics from VI images and integrate them into IR images. Since the existing unsupervised GAN-based image translation approaches face several challenges, such as incorrect mapping and lack of fine details, we propose a structure connection module and a region-of-interest (ROI) focal loss method to address the current limitations. Experimental results show the superiority of the IR2VI algorithm over baseline methods. 
\end{abstract}


Humans have poor night vision compared to many animals, partly because the human eye lacks a tapetum lucidum \cite{chijiiwa1990histological}. This biological deficiency may lead to several undesirable fatalities. For example, vehicle collisions are much more likely to happen at night than during daytime. \cite{sullivan1999assessing}. Hence, context enhancement plays a critical role in many night vision applications.

A straightforward way to enhance the context in night vision is by employing thermal or infrared (IR) and visible (VI) image fusion approaches \cite{bhatnagar2015novel,ulhaq2016face, zhou2016fusion}, where an IR sensor can enhance thermal objects in a night environment from a visual spectrum background \cite{liu2012objective}. However, an IR/VI image fusion method only works at dawn or dusk when the visible camera is still able to capture the visual scene. When there is a dark night without any sort of moon or artificial lights, only the IR sensor works. Technically, the emitted energy of an object reaches the IR sensor which can be converted into a temperature value, and thus the IR sensor can see in the night. However, IR image lacks fine semantic information as textures seldom influence the heat emitted by an object. When the image is presented to a final user, a visible image is preferred because it is more suitable to the sensitivity of human visual perception system ranging from $400nm$ to $700nm$.  
 
In a nighttime scenario, translating an IR images to a VI image would be a possible solution to enhance environmental perception at night. In recent years, numerous research has been proposed to solve this challenging task by colorizing the IR images using different models \cite{limmer2016infrared,suarez2017learning,suarez2017infrared}. Recent progress in machine learning might advance nighttime imagery. Generally,  machine learning models are often employed to predict the color values directly. However, those models need large-scale datasets with corresponding ground truth data for training. For the IR image captured at night, it is almost impossible to find a pixel-to-pixel aligned day-time to the VI image. In addition, the semantic information from the visible spectrum comes with texture and structure, as well as color.

We can formulate the task of translating the night-time IR images to the day-time VI images as an unsupervised image-to-image translation problem, where we aim to model the mapping between the two different data distributions without fully paired training datasets. This is a significant challenge until the Generative Adversarial Networks (GANs) based methods were proposed \cite{star,goodfellow2014generative,unit,cycle} in the most recent years. The basic idea behind these methods is that a generative Convolutional Neural Network (CNN) can translate an image from the source data domain to the target data domain, while a discriminative CNN can distinguish the translated image from the real image. The generator attempts to fool the discriminator and the discriminator attempts to identify the image from the generator as fake. In this way, the GANs will end up in local Nash Equilibrium. However, when applying this unsupervised image-to-image translation framework to the IR-to-VI task directly, two major problems arise. Firstly, the trained models face an incorrect mapping problem, when most of areas from the input image are overly bright. Secondly, the generated image lacks fine details, especially for the small objects.

To address above mentioned problems, we proposed an unsupervised IR-to-VI image translation framework, namely IR2VI. Basically, IR2VI is a GAN-based method, and the basic architecture is comprised of one generator and two discriminators, i.e., a global discriminator and a Region-of-Interest (ROI) discriminator. To deal with the incorrect mapping problem, we added a structure connection in the generator enabling the generated image to keep original structure information. Moreover, we also proposed the ROI focal loss which consists of an  ROI cycle-consistency loss and an ROI adversarial loss to resolve more fine details in the concerned areas. To summarize, the contributions of this paper include:
\begin{itemize}
\item A novel unsupervised thermal image translation framework, IR2VI, is proposed to enhance the environmental perception at night by translating night-time IR images to day-time VI images.
\item A structure connection and an ROI focal loss are implemented to deal with the existing problems with GAN-based methods, e.g., incorrect mapping.
\end{itemize}
Both subjective and quantitative results are given in the experiments, which demonstrate the superiority of IR2VI over the baseline models.

\section{Related Work}
\subsection{Infrared and Visible Image Fusion}
IR and VI image fusion is an active research in the last two decades, where the objective is to fuse the IR and the VI image into a composite image to boost imaging quality for improved visual capability of human and robot machines \cite{jin2017survey}. The image fusion methods can be roughly categorized into methods in spatial domain and transform domain. The implementation in the spatial domain is straightforward, such as weighted average and gradient transfer fusion \cite{ma2016infrared}. The transform-domain based algorithms include non-subsampled contourlet transform (NSCT) \cite{bhatnagar2015novel}, wavelet \cite{shamsafar2014fusing}, guided filter \cite{zhou2016fusion}, etc. These transform image fusion methods are developed with the assumption that the IR and VI images are fully registered. Nevertheless, the visible camera does not function in most night environments, which means only the IR image can be acquired and the image fusion operation cannot be further performed.

\subsection{Infrared Image Colorization}
IR image colorization is a type of color transferring technique which aims at transforming a gray-scale IR image into a multi-channel RGB image. Basically, this technique can be divided into non-parametric and parametric based methods. The non-parametric based methods \cite{gupta2012image,hamam2012single,eric_spie} generally require colorful reference images whose structure is also similar to the source IR image, and then the methods utilize the image analogies framework \cite{hertzmann2001image} to transfer the color onto the IR image. While the parametric based methods \cite{limmer2016infrared,suarez2017learning,suarez2017infrared} can directly estimate chrominance values by training one or multiple prediction models, such as deep convolutional neural networks (DCNNs) \cite{limmer2016infrared} or GANs \cite{suarez2017learning,suarez2017infrared}. However, these colorization approaches either require paired pixel-wise aligned training dataset or rely on a colorful reference image, which is hardly acquired in a night vision application. Contrasted with IR image colorization methods, our IR2VI can mapping the intrinsic features from VI image to the IR image and does not need a fully registered dataset.

\subsection{Image-to-Image Translation}
Image-to-Image translation is to learn a mapping function from a source data distribution to one or multiple data distributions. Recent progresses in this field were achieved with GANs \cite{goodfellow2014generative}. These GAN approaches can be categorized into supervised and unsupervised ones. For the supervised models \cite{isola2017image,wang2017high}, the L1 loss function is commonly adopted and thus the paired images are required. While the unsupervised models \cite{star,unit,cycle} alleviate the difficulty for obtaining data pairs with different techniques, such as variational auto-encoders (VAEs) \cite{unit} or cycle consistency \cite{cycle}. However, the unsupervised methods can also lead to several undesirable problems, such as incorrect mappings, when applied to the IR-to-VI image translation task. In our IR2VI framework, we designed a structure connection module and ROI focal loss to successfully address these problems.

\section{The IR2VI Framework}
\subsection{Overall Architecture}
\begin{figure*}[!h]
\centering
\includegraphics[width=0.95\textwidth]{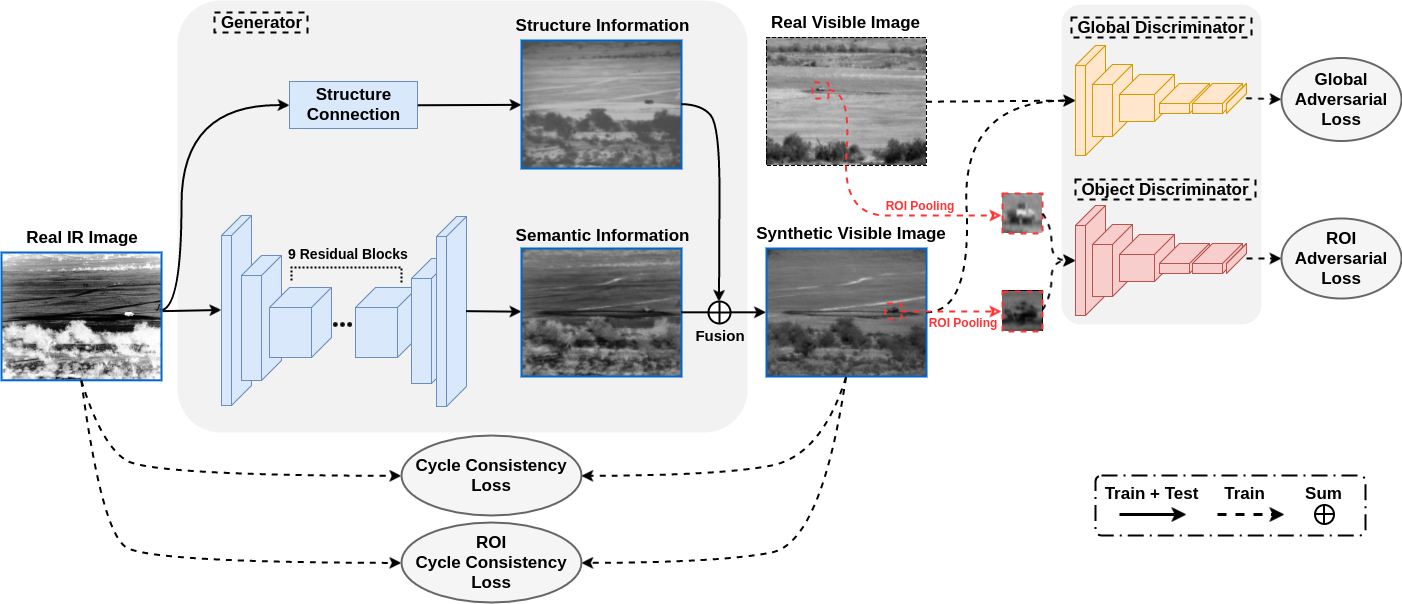}
\caption{An overall architecture of the proposed IR2VI framework. Note that this is a brief illustration of the architecture, which actually needs to be duplicated for training in the CycleGAN way. }
\label{fig:pipeline}
\end{figure*}

As we can see in the Fig.~\ref{fig:pipeline}, the basic architecture of IR2VI includes a generator, a global discriminator, and an ROI discriminator. The generator translates an IR image to a synthetic VI image that looks similar to the real VI image, while the global discriminator distinguishes translated VI images from real ones. The ROI discriminator aims to distinguish the ROIs between translated VI image and real ones. In this way, the synthetic VI images are designed to be indistinguishable from the real VI images.

Similar to CycleGAN \cite{cycle} and StarGAN \cite{star}, we adopted the residual auto-encoder architecture from Johnson et al. \cite{johnson2016perceptual} with 9 residual blocks \cite{he2016deep} for the generative network. We follow the naming convention used in image translation community \cite{unit,wang2017high,cycle}, with the network configuration expressed as follows:
\begin{align*}
c7s1{-}32, d64, d128, R128, R128, R128, \\
R128, R128, R128, R128, R128, R128, \\
u64, u32, c7s1{-}1, c7s1{-}1^{structure}, F
\end{align*}
where the $c7s1{-}k$ represents a $7\times 7$ Convolution-BatchNorm-ReLU layer with $k$ filters and stride $1$. And the right top $structure$ means that is for structure connection module which will be introduced in the following subsection. $dk$ denotes a $3 \times 3$ Convolution-BatchNorm-ReLU (CBR) layer with $k$ filters, and stride 2. We also employed reflection padding to reduce boundary artifacts. $Rk$ denotes a residual block which consists of two $3 \times 3$ convolutional layers with the same number of filters on both layer. $uk$ represents a $3 \times 3$ fractional strided CBR layer with $k$ filters, and stride $\frac{1}{2}$. $F$ denotes fusion layer where we utilize $sum$ and $tanh$ functions to fuse the output information from both structure connection and residual auto-encoder. We adopted the PatchGAN \cite{isola2017image} with 4 hidden layers for all the discriminative networks, with the network configuration is as follows: 
\begin{align*}
C64, C128, C256, C512, C512
\end{align*}
where $Ck$ denotes a $4 \times 4$ Convolution-BatchNorm-LeakyReLU layer with $k$ filters and stride $2$ (except for the last layer with stride $1$). After the last layer, we applied a convolution to produce a $1$ dimensional output. BatchNorm is not applied to the first $C64$ layer. We set the slope $0.2$ for leakyReLU.

For training the IR2VI, four loss functions were utilized (cycle consistency loss, global adversarial loss, ROI cycle-consistency loss, and ROI adversarial loss). Details about each loss function are provided in the following sections.

Basically, the IR2VI framework evolves from the CycleGAN \cite{cycle}. In contrast to CycleGAN, we made two important improvements: (1) A structure connection module has been added into the generator to constrain the structure deformation; and (2) a ROI focal loss is calculated in the training stage, which enables the critical regions to be focused in translation procedure.  

\subsection{Implementation Details}
\subsubsection{Structure Connection}

\begin{figure}[!h]
\centering
\includegraphics[width=\linewidth]{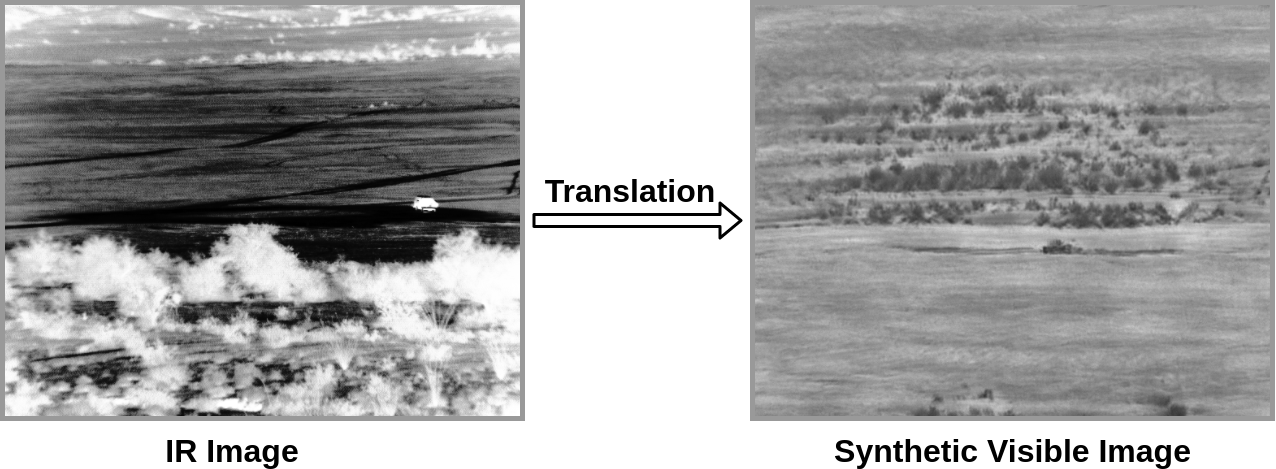}
\caption{An example of the results from the CycleGAN to illustrate the incorrect mapping problem.}
\label{fig:wrong_mapping}
\end{figure}

Incorrect mapping is a common issue for the unsupervised image translation models which directly lack supervised signals. When objects in the source image are overly bright, which is an extremely common situation for the IR image at night, the translation models will be confused and map the objects to any random permutation of objects in the target domain. As the example in Fig.~\ref{fig:wrong_mapping}, where the CycleGAN wrongly mapped the ground to the forest and the vehicle to a different object. To solve the incorrect mapping problem, we added a shortcut to the generator to connect input image with generated image, which is called ``structure connection." A $7\times7$ convolution layer is adopted to extract the detailed structure information from the IR image and then fuse it with the semantic information generated by the residual auto-encoder model. In this way, the deep CNN is able to focus on the semantic level task while the structure connection enables the synthetic VI image to keep original structure information.

\subsubsection{Cycle Consistency Loss}
The cycle consistency loss was proposed by Zhu et al. in \cite{cycle}. The basic idea is to learn two mappings $G: \mathrm{IR} \to \mathrm{VI}$, and $F: \mathrm{VI} \to \mathrm{IR}$, which can translate the image between two domains. For the $x \in \mathrm{IR}$, it forces $F(G(x)) \approx x $, while for $y \in \mathrm{VI}$, it forces $G(F(y)) \approx y$. Thus, it becomes possible to constrain the cycle-consistency and eliminate undesirable mappings. The cycle consistency loss can be formulated as follows:
\begingroup\makeatletter\def\f@size{7}\check@mathfonts
\def\maketag@@@#1{\hbox{\m@th\normalsize\normalfont#1}}%
\begin{equation}
\begin{aligned}
\mathcal{L}_{cyc}(G,F) = & \mathbb{E}_{x \sim \mathrm{IR}_{data}}[\lVert F(G(x)) -x {\rVert}_1] \\
& + \mathbb{E}_{y \sim \mathrm{VI}_{data}}[\lVert G(F(y)) -y {\rVert}_1].				
\end{aligned}
\end{equation} \endgroup
\subsubsection{Global Adversarial Loss}
The global adversarial loss is derived from the global discriminator, which aims to distinguish the full-size image from the real domain with the full-size image from the synthetic domain. Because the image fed to the discriminator network is full-sized, the adversarial loss is designated as the global adversarial loss. As aforementioned, two mapping functions are created to manipulate the cycle consistency loss. The global adversarial loss is applied to both mapping functions. Taking the mapping function $G: \mathrm{IR} \to \mathrm{VI}$ as example, its discriminator is $D^{G}_{\mathrm{VI}}$. Thus, the global adversarial loss is formulated as:
\begingroup\makeatletter\def\f@size{7}\check@mathfonts
 \def\maketag@@@#1{\hbox{\m@th\normalsize\normalfont#1}}%
\begin{equation}
\begin{aligned}
\mathcal{L}_{adv}(G,D^{\mathbf{g}}_{\textrm{VI}}, \mathrm{IR}, \mathrm{VI}) =  \mathbb{E}_{y \sim \mathrm{VI}_{data}} \left[\log ({D^{\mathbf{g}}_{\mathrm{VI}}(y)}) \right] \\
 + \mathbb{E}_{x \sim \textrm{IR}_{data}} \left[\log (1-{D^{\mathbf{g}}_{\mathrm{VI}}(G(x))}) \right]
\end{aligned}
\end{equation} \endgroup

\subsubsection{ROI Focal Loss}
Generally, the generated images via adversarial training are often lack of fine details and realistic textures \cite{chen2017photographic,wang2017high}. This is manifested when the concerned object is extremely small. To end this, we propose a region of interest (ROI) focal loss which consists of ROI adversarial loss and ROI cycle-consistency loss. The ROI approach is suitable for those training dataset with bounding boxes. In contrast to the cycle consistency loss and global adversarial loss which take the full-size image as input, the ROI focal loss operates in the ROI. To obtain the ROIs from the full-size image, the ROI pooling layer \cite{fast-rcnn} is adopted, which was proposed to solve the object detection challenge. Based on provided bounding boxes, the ROI pooling layer is able to crop and reshape the arbitrary area to the fixed size image. In our work, we set $64 \times 64$ as the fixed size of ROI image and name the ROI pooling function as $R(\cdot)$. Same as the cycle consistency loss and global adversarial loss, the ROI focal loss is applied to both mapping functions. Here, the mapping function $G: \mathrm{IR} \to \mathrm{VI}$ is used as an example.

The ROI cycle-consistency loss can be formulated as follows:
\begingroup\makeatletter\def\f@size{7}\check@mathfonts
\def\maketag@@@#1{\hbox{\m@th\normalsize\normalfont#1}}%
\begin{equation}
\begin{aligned}
\mathcal{L}^{\mathbf{roi}}_{cyc}(G,F) =  \mathbb{E}_{x \sim \mathrm{IR}_{data}}[\lVert R(F(G(x))) -R(x) {\rVert}_1] \\
 + \mathbb{E}_{y \sim \mathrm{VI}_{data}}[\lVert R(G(F(y))) -R(y) {\rVert}_1].				
\end{aligned}
\end{equation} \endgroup

The network configuration of ROI discriminator is same as that of the global discriminator. The ROI adversarial loss can be expressed as follows:
\begingroup\makeatletter\def\f@size{7}\check@mathfonts
\def\maketag@@@#1{\hbox{\m@th\normalsize\normalfont#1}}%
\begin{equation}
\begin{aligned}
\mathcal{L}^{\mathbf{roi}}_{adv}(G,D^{\mathbf{roi}}_{\mathrm{VI}}, \mathrm{IR}, \mathrm{VI}) = \mathbb{E}_{y \sim \mathrm{VI}_{data}} [\log ({D^{\mathbf{roi}}_{\mathrm{VI}}(R(y))})] \\
+ \mathbb{E}_{x \sim \mathrm{IR}_{data}}[\log (1-{D^{\mathbf{roi}}_{\mathrm{VI}}(R(G(x)))})], 
\end{aligned}
\end{equation} \endgroup
where $D^{roi}_{\mathrm{VI}}$ represents the ROI discriminator for VI images.

\subsection{Full Objective}
Finally, the complete objective function can be written as: 
\begingroup\makeatletter\def\f@size{7}\check@mathfonts
\def\maketag@@@#1{\hbox{\m@th\normalsize\normalfont#1}}%
\begin{equation}
\begin{aligned}
\mathcal{L}_{full}= & \mathcal{L}_{adv}(G,D^{\mathbf{g}}_{\mathrm{VI}}, \mathrm{IR}, \mathrm{VI}) 
+ \mathcal{L}_{adv}(G,D^{\mathbf{g}}_{\mathrm{IR}}, \mathrm{IR}, \mathrm{VI}) \\
& + \lambda_{cyc}\mathcal{L}_{cyc}(G,F) 
+ \lambda_{roi}({\lambda_{cyc}}\mathcal{L}^{\mathbf{roi}}_{cyc}(G,F) \\
&+ \mathcal{L}^{\mathbf{roi}}_{adv}(G,D^{\mathbf{roi}}_{\mathrm{VI}}, \mathrm{IR}, \mathrm{VI})
+ \mathcal{L}^{\mathbf{roi}}_{adv}(G,D^{\mathbf{roi}}_{\mathrm{IR}}, \mathrm{IR}, \mathrm{VI})),
\label{loss:full_loss}
\end{aligned}
\end{equation}
\endgroup
where $\lambda_{cyc}$ and $\lambda_{roi}$ are the hyper-parameters that control the relative importance of cycle consistency loss and the ROI focal loss. For simplicity, 
$\mathcal{L}_{full}$ represents $\mathcal{L}(G,F,D^{\mathbf{roi}}_{\mathrm{VI}},D^{\mathbf{roi}}_{\mathrm{IR}},D^{\mathbf{g}}_{\mathrm{VI}},D^{\mathbf{g}}_{\mathrm{IR}}) $. Finally, the method resolves:

\begingroup\makeatletter\def\f@size{7}\check@mathfonts
\def\maketag@@@#1{\hbox{\m@th\large\normalfont#1}}%
\begin{equation}
\begin{aligned}
G^*,F^* = \arg \min_{F,G} \max_{D^{\mathbf{roi}}_{\mathrm{VI}},D^{\mathbf{roi}}_{\mathrm{IR}},D^{\mathbf{g}}_{\mathrm{VI}},D^{\mathbf{g}}_{\mathrm{IR}}}\mathcal{L}_{full}
\end{aligned}
\end{equation} 
\endgroup



\subsection{Evaluation Protocol}
\label{section:protocol}
As there is no ground truth associated with the translated image, it is hard to evaluate the performance of the different image translation methods. In this study, we focused on the dataset with bounding boxes. Thus, it is possible to assess different methods by performing object detection on the synthesized images. Specifically, We adopted the object detector of Faster R-CNN with ResNet 101 network presented in \cite{huang2017speed}, and trained it on the day-time VI image dataset (target domain). Then, different image translation methods were empoyed to generate the synthetic VI image from the IR image at night. Finally, the trained object detector model was performed on the synthetic VI image collections. One choice is to use the de-facto standard average precision (AP) to evaluate the performance of object detector, which is calculated as the ratio between the area under Precision-Recall curve (less than 1) to the entire area (which is 1).

\begin{figure*}[!h]
\centering
\includegraphics[width=0.95\textwidth]{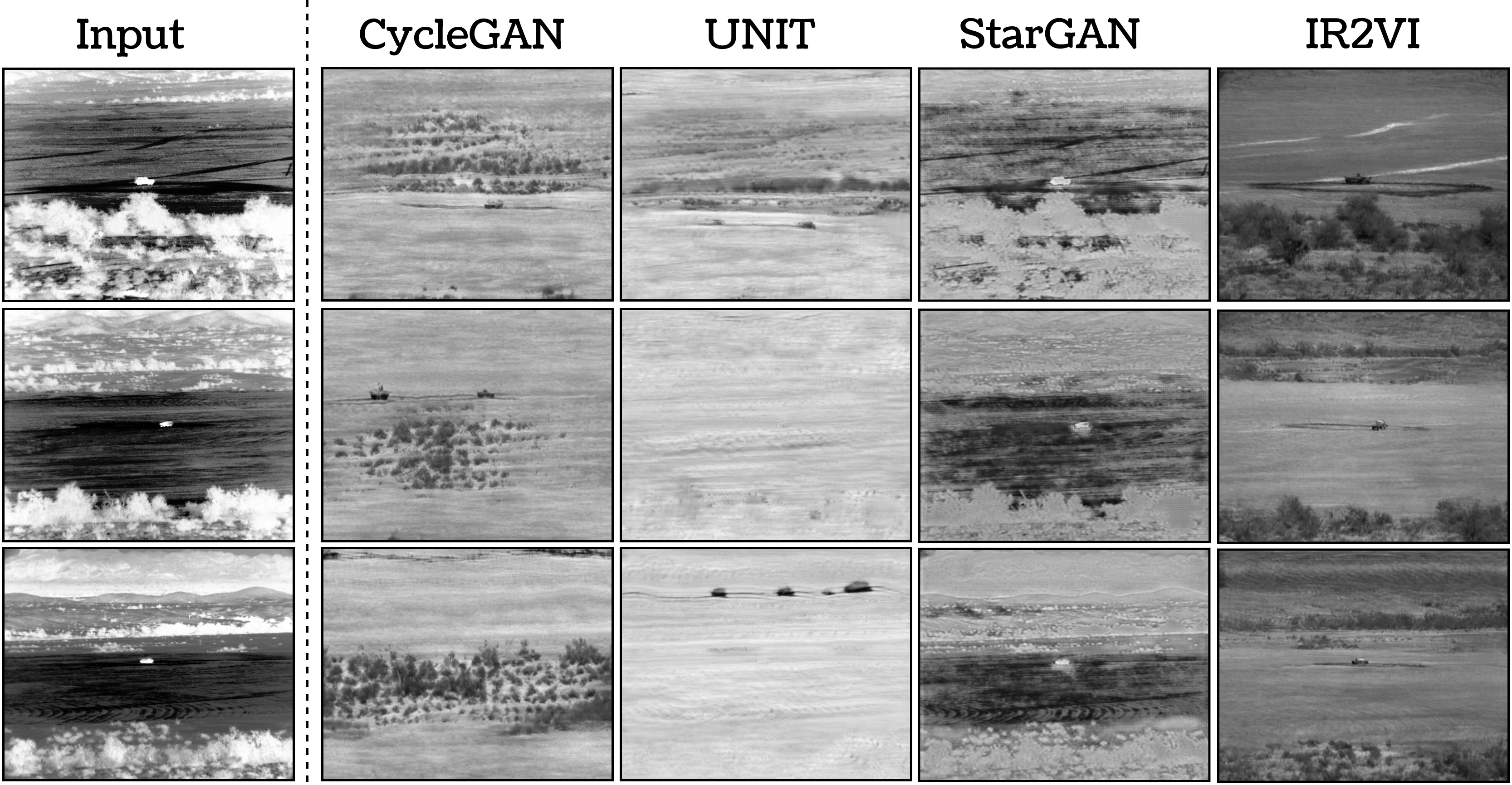}
\caption{Comparison of different results on the SENSIAC dataset.}
\label{fig:visual_comparison}
\end{figure*}

\section{Experimental Results}
This section introduces the Military Sensing Information Analysis Center (SENSIAC) dataset that is used in all the experiments. Then the settings of the hardware and training detail of IR2VI are listed. Lastly, the IR2VI is compared against state-of-the-art methods for subjective qualitative and objective quantitative analysis.

\subsection{Dataset}
SENSIAC \cite{SENSIAC} recently released a large-scale military IR-VI image dataset for automatic target recognition (ATR). In this study, the proposed IR2VI framework is evaluated with the SENSIAC dataset along with the state-of-the-art methods. Basically, the SENSIAC dataset contains 207GB of middle-wave infrared (MWIR) videos and 106GB of VI (grayscale) videos along with manually labeled bounding boxes. Various types of objects are included in this dataset, for instance, soldiers, military vehicles, and civilian vehicles. It worth noting that the dataset was collected during both day-time and night-time with multiple observation distances from 500 to 5000 meters. However, it has paired IR-VI videos in day-time but only has IR videos in the night-time.

The objective of this study is to translate the night-time thermal images to the day-time VI images, where only the night-time IR videos and the day-time VI videos are used in the experiments. We selected 3 different observation distances, e.g., 1000, 1500, and 2000 meters, and split into training/testing datasets \cite{shuo}. For training the image translation models, we sampled the keyframe at 3$Hz$ (every 10 frames). Thus, there are 2700 night-time IR training images and 2691 day-time VI training images. Note that all the night-time IR images are preprocessed by histogram equalization operation prior to being fed into the models. For training the object detector, the keyframe is sampled at 6$Hz$ (every 5 frames). So, there are 4573 day-time VI images and 5400 night-time IR images in training dataset. Meanwhile, there are 2812 day-time VI images and 3240 night-time IR images in testing dataset.

\subsection{Experimental Setup}
The IR2VI was developed based on CycleGAN \cite{cycle} by using Pytorch deep learning toolbox~\cite{pytorch}. We used a workstation with an NVIDIA GeForce GTX 1080 GPU, an Intel Core i7 CPU and 32 GB Memory. 

For the hyper-parameters, the parameters are $\lambda_{cyc} = 5$ and $\lambda_{roi} = 0.1$ in Equation \ref{loss:full_loss}. All the networks were trained from scratch, and the weights were initialized from a Gaussian distribution with zero mean and $0.02$ standard deviation. The Adam solver was employed with a batch size of $2$ and set the learning rate at $0.0002$ for the first 20 epochs and a linearly decaying rate to zero over the next 20 epochs. 

For the fair comparison, we did not modify the default setting of the baseline methods except the image channel, image size, batch size, and training epochs. To be specific, the images in SENSIAC dataset are grayscale, so the input channel was set to $1$ for the input channel of all the networks and the output channel of the generation network. Because the limited capacity of the GPU memory, the training epochs were set to $40$ with batch size $2$ for CycleGAN \cite{cycle} and UNIT \cite{unit}, training epochs $40$ with batch size $12$ for StarGAN \cite{star}. And the images were center-cropped to $256 \times 256$ pixels before feeding into the baseline networks. Our IR2VI is a kind of object-based framework, so the images were cropped to $256 \times 256$ with at least one object. Because the generator network of every method is a fully CNN which is able to take an image of arbitrary size as input, the full-size image is fed to the network in the testing stage.

\subsection{Results}
In this section, we compared with the state-of-the-art unsupervised image translation methods: CycleGAN \cite{cycle}, UNIT \cite{unit} and StarGAN \cite{star}.

\subsubsection{Subjective Comparisons}
All the methods were trained on the same training set and tested on the unseen images. Figure \ref{fig:visual_comparison} shows the translated images from unseen images by different methods. It is apparent that the CycleGAN and the UNIT have the serious incorrect mapping problems. The CycleGAN could not tell where are the trees and ground, so it mapped the ground to a forest. In the second testing image, the CycleGAN incorrectly generated two vehicles. The translated images by UNIT are almost similar without too much semantic information. For the StarGAN method, it has few incorrect mapping problems but lacks sharp texture information. Significantly, we qualitatively observed that our IR2VI provided the highest visual quality of translation results compared to the baseline methods. It can not only bring the spatial semantic information but also makes the target clear. We believe that our IR2VI framework benefits from the advantages of the structure connection module and the ROI focal loss.

\subsubsection{Quantitative Comparisons}

\begin{figure}[!h]
\centering
\includegraphics[width=0.9\linewidth]{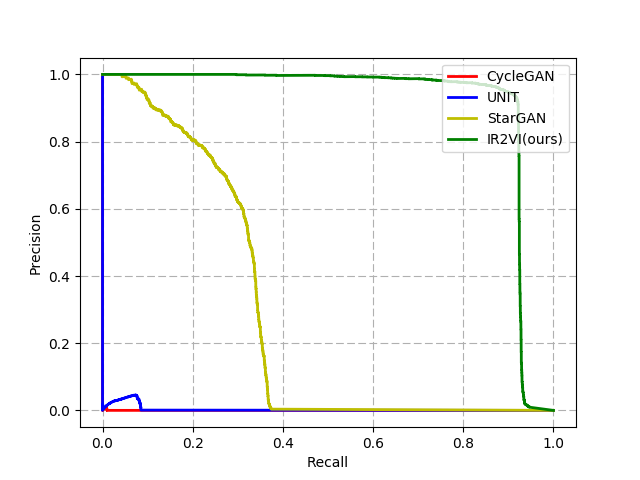}
\caption{Precision-Recall curve of the object detector on different synthesis images.}
\label{fig:AP}
\end{figure}

For the quantitative objective evaluation, we applied the evaluation protocol introduced in Section \ref{section:protocol}. Figure \ref{fig:AP} and Table \ref{tab:AP_score} show the Precision-Recall (PR) curves and Average Precision (AP) scores of the object detector on translated images by different translation methods.

\begin{table}[!htpb]
\caption{Average precision scores of the object detector on the generated testing images by different translation methods.}
\label{tab:AP_score}
\centering
\resizebox{\linewidth}{!}
{
\begin{tabular}{@{}crrrcrrrcrrrcrrrcrrrc@{}}\toprule &&
 \multicolumn{1}{c}{\textbf{CycleGAN}} & &
 \multicolumn{1}{c}{\textbf{UNIT}}  & &
 \multicolumn{1}{c}{\textbf{StarGAN}}  & &
 \multicolumn{1}{c}{\textbf{IR2VI}}   \\ \midrule

\textbf{AP (\%)} && $7.62\times10^{-3} $&& $0.37$ && $28.48$ && \textbf{91.70}  \\
\bottomrule
\end{tabular}
}
\end{table}

The results clearly show that there is a large margin between different methods, and our IR2VI achieved the best AP score at $91.70 \%$ which has a $63.22\%$ margin to the second rank method, StarGAN. These results demonstrate that the IR2VI is capable of adding semantic visible information and also add object shape information to the original thermal images. Even though the translated images by the StarGAN lack texture information, the blur shape information can also help the VI object detector to accomplish detection. However, the incorrect mapping problems in CycleGAN and UNIT made the VI detector completely fail as indicated with a nearly zero AP score.


\section{Conclusion}
In this paper, we proposed an unsupervised thermal image translation framework for context enhancement at night, called IR2VI. Thanks to the proposed structure connection module in the generative network, IR2VI is able to overcome the incorrect mapping problem which is commonly faced by the state-of-the-art image translation methods. Moreover, the proposed ROI focal loss enables IR2VI to generate a synthetic VI image with more fine details as compared with baseline models. The results demonstrate the IR2VI contributions which not only broaden the area of context enhancement for night vision but also can be applied to many other related research fields within image fusion.

\section*{Acknowledgements}
The authors would like to thank NVIDIA for supporting our research by offering computing equipments. And this research was enabled in part by the support from West Grid \href{https://www.westgrid.ca/}{ (www.westgrid.ca) } and Compute Canada \href{https://www.computecanada.ca/}{ (www.computecanada.ca) }. In particular, the authors express their sincere gratitude to Huan Liu and Junchi Bin for the helpful discussions when this work was being carried out. 
{\small
\bibliographystyle{ieee}
\bibliography{egbib}
}

\end{document}